\documentclass[fleqn,10pt]{wlscirep}
\usepackage[utf8]{inputenc}
\usepackage[T1]{fontenc}
\usepackage{multirow}
\usepackage{svg}
\usepackage[font=sf]{caption}

\usepackage[square,comma,numbers,sort&compress]{natbib}
\usepackage{xcolor}
\usepackage{colortbl}

\definecolor{hellgrau}{rgb}{0.95,0.95,0.95}

\title{Deep Transfer-Learning for patient specific model re-calibration: Application to sEMG-Classification}

\author[1,3,*]{Stephan Johann Lehmler}
\author[1]{Muhammad Saif-ur-Rehman}
\author[2]{Tobias Glasmachers}
\author[1]{Ioannis Iossifidis}
\affil[1]{Institute of Computer Science, University of Applied Science Ruhr West, Mülheim an der Ruhr, Germany}
\affil[2]{Institute for Neural Computation, Ruhr University, Bochum, Germany}
\affil[3]{Faculty of Electrical Engineering and Information Technology, Ruhr-University, Bochum, Germany}

\affil[*]{stephan.lehmler@hs-ruhrwest.de}
\affil[+]{Muhammad.Saif-ur-Rehmann@hs-ruhrwest.de}
\affil[+]{tobias.glasmachers@ini.rub.de}
\affil[+]{iossifidis@hs-ruhrwest.de}

\begin{abstract}
Accurate decoding of surface electromyography (sEMG) is pivotal  for muscle-to-machine-interfaces (MMI) and their application for e.g. rehabilitation therapy.  sEMG signals have high inter-subject variability, due to various factors, including skin thickness, body fat percentage, and electrode placement. Therefore, obtaining high generalization quality of a trained sEMG decoder is quite challenging.   Usually, machine learning based sEMG decoders are either trained on subject-specific data, or at least recalibrated for  each user, individually. Even though, deep learning algorithms produced several state of the art results for sEMG decoding,however, due to the limited amount of availability of sEMG data, the deep learning models are prone to overfitting. The requirement of long training time is another reason for the limited use of deep learning algorithms in MMI decoding applications.  
Recently, transfer learning for domain adaptation improved generalization quality with reduced training time on various machine learning tasks. In this study, we investigate the effectiveness of transfer learning using weight initialization for recalibration of two different pretrained deep learning models on a new subjects data, and compare their performance to subject-specific models. To the best of our knowledge, this is the first study that  thoroughly investigated weight-initialization based transfer learning for sEMG classification and compared transfer learning to subject-specific modeling.
We tested our models on three publicly available databases under various settings. On average over all settings, our transfer learning approach improves 5 \%-points on the pretrained models without fine-tuning and  12 \%-points on the subject-specific models, while being trained on average 22\% fewer epochs. Our results indicate that transfer learning enables faster training  on fewer samples than user-specific models, and improves the performance of pretrained models as long as enough data is available. 
\end{abstract}
\begin{document}

\flushbottom
\maketitle
%
%

\section*{Introduction}
Reliable decoding of intended movements using sEMG signals is crucial for the restoration of movements of disabled persons. Henceforth, development of a reliable and robust MMI decoder remains one of the primary goals of the BMI-community. Real life applications, such as control of prosthesis, rehabilitation-aiding exoskeletons, and MMI for gaming are active  research areas. Recently published studies on movement classification based on sEMG showed promising results with high accuracy and real-time processing capabilities \cite{8512820, 8911244, faust2018deep}.  However,  the presence of intra- and interpersonal variability in the  sEMG signals can still complicate the calibration of a model to a new user.\\
This paper investigates  the effects of interpersonal variability on the classification accuracy of deep learning based sEMG classifiers.\\
In previously published studies, deep Neural Networks showed promising classification accuracy of sEMG signals at inference time, and outperformed conventional  statistical and machine learning models, especially on more complex classification tasks with less manual feature engineering \cite{faust2018deep, tao2019multi, xiong2021deep}.
Even though the results of recent studies settings seem to be very promising in lab settings, real life applications are much more complicated because the adaptation to unseen users is quite challenging.\\
The sEMG signals exhibit high variance between subjects \cite{araujo2000inter}, which is  increased between healthy and amputated subjects \cite{campbell2020comparison, campbell2019differences}.
Other factors such as electrode placement \cite{hogrel1998variability}, fatigue \cite{linssen1993variability}, heat \cite{herrera2020temperature}, skin fat ratio \cite{nordander2003influence} and gender \cite{meduri2016inter} also influence the inter- and intra-subject variance of sEMG measurements. This high inter-subject variability challenges the machine learning models to adapt high generalization quality on unseen subjects.\\
The effect of this variability has been studied to some extent for other aspects of the signal processing pipeline (e.g. normalization \cite{burden2010should}) or the classification algorithms (e.g. Random Forests \cite{palermo2017repeatability}, SVM \cite{kanoga2021semi, 6977190} or Muscle Source Activation Models \cite{kim2020subject}).\\
While simple models can relatively easily be re-trained on each new user, this is more challenging for larger neural networks. This is due to their huge parameter space, which result in comparatively long training times and a high-risk of overfitting on small samples sizes. Therefore, implications of deep learning  in a rehabilitation framework is limited, where short calibration using only a few repetitions of each patient are desirable.
So far, however, there is a lack in research on the effects of interpersonal differences on the performance of neural network based sEMG classifiers.\\
Some recent works addressed the problem using few shot learning \cite{rahimian2021fs}, adaptation layers with RNNs \cite{ketyko2019domain} and adaptive batch normalization on CNNs \cite{lin2020normalisation, cote2017transfer, du2017surface} \\
In this paper, we  thoroughly evaluated the effectiveness of basic weight-initialization based transfer learning for user adaptation and compared the performance and training times of user-specific deep learning models. To the best of our knowledge,  no one has previously investigated the weight-initialization based transfer learning for EMG based decoder \\
Possible benefits of weight-initialization based transfer learning are the straight-forward implementation and higher flexibility during model building, because fine-tuning on new user data can in principle be done with any pretrained model.
This paper thoroughly investigates how well this basic form of transfer-learning is suited for the task of sEMG classification and how much knowledge transfer takes place.
Most papers so far only compared performance of classifiers with and without transfer learning in fixed settings. In this paper, we compare subject-specific, pretrained and fine tuned models under varying conditions.We aim to gain insights into the question under which circumstances transfer learning based user-calibration of sEMG classifiers performs well.

\section*{Method}
\subsection*{Transfer Learning}

\begin{figure}[htp]
\centering
\includegraphics[width=.8\hsize]{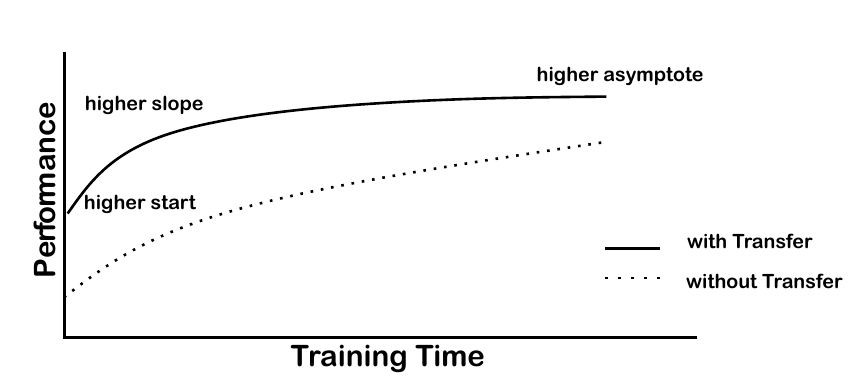}
\caption{Three possible signs of successful knowledge transfer}
\label{fig:transfer}
\end{figure}

Generally, transfer learning denotes machine learning applications where knowledge learned in one situation is transferred to a different but related situation. Domain adaptation is a more specific term, referring to the case where the task (input-output-mapping) stays the same, but the input or output data distribution is slightly different at the training and inference time, usually necessitating a slight change of the model parameters. In this case, applying a pretrained deep-learning model for sEMG classification to a new user would be an example of domain adaptation.\\
There are several  approaches for transferring knowledge to a new domain. In this paper, we utilize the commonly used approach of weight-initialization and subsequent fine-tuning. For calibrating our deep learning models to a new user, we transfer the weights from a pretrained model instead of random initialization and subsequently fine-tune it on data from the new user.\\ 
The effect of transfer learning can be seen by comparing the model performance during training with and without transfer learning. Effective knowledge transfer can be noticed in either or a higher starting performance, a higher slope (faster learning) and an higher asymptotic performance.\\
We compare deep learning models for sEMG classification trained and evaluated on only one subject (subject-specific models) with transfer learning models, where we pretrain the model on other users data and further fine tune the weights on the subject used for evaluation. If transfer learning takes places we expect similar curves to figure \ref{fig:transfer}.

\subsection*{Data}
\begin{figure}[htp]
\centering
\begin{minipage}{0.45\textwidth}
\includegraphics[width=\hsize]{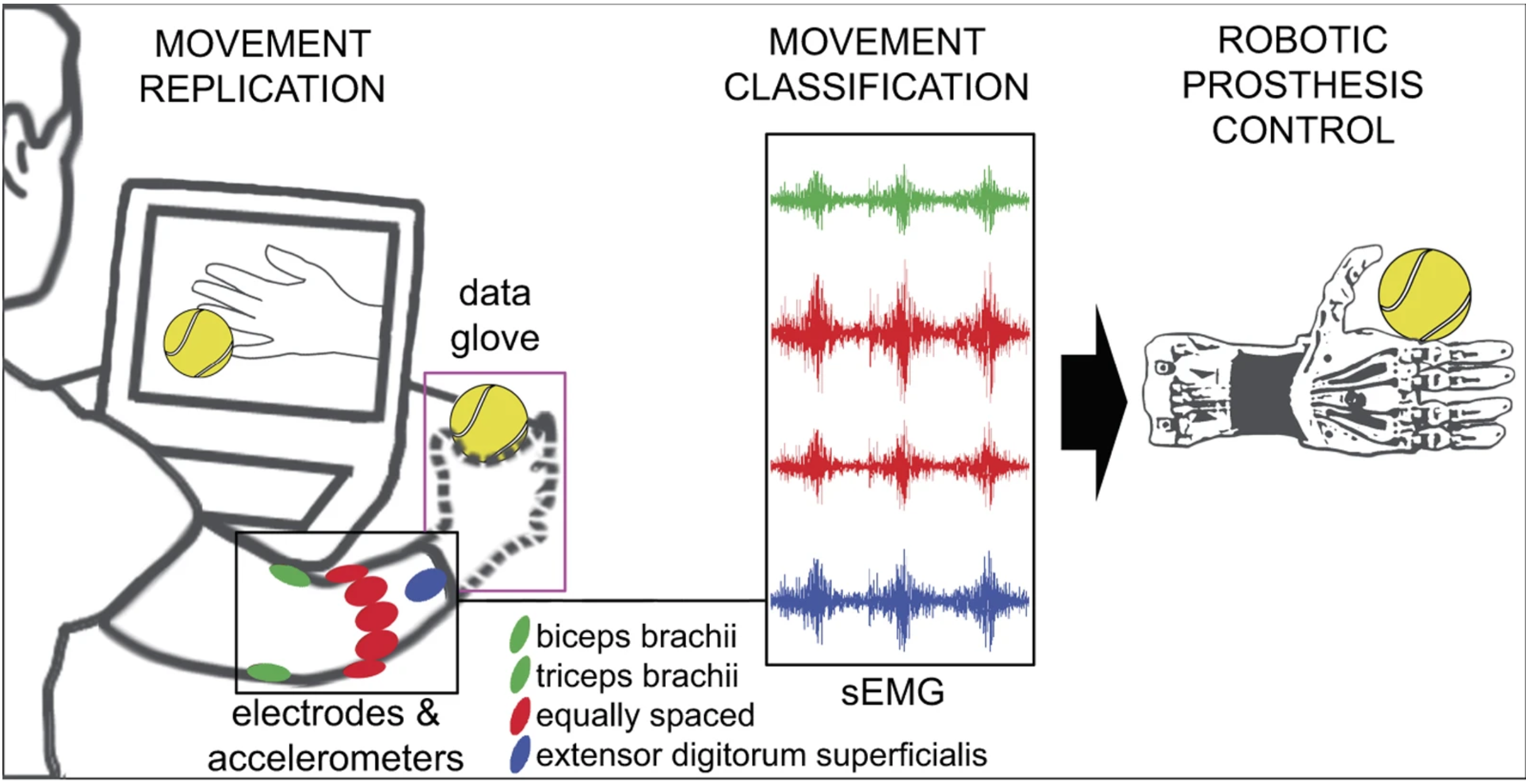}
\end{minipage}
\begin{minipage}{0.45\textwidth}
\includegraphics[width=\hsize]{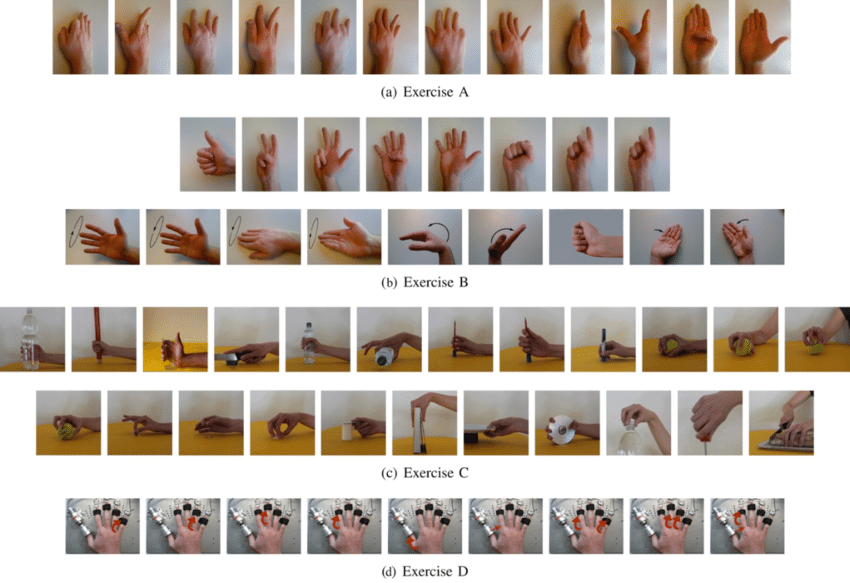}
\end{minipage}
\caption{Experimental setup and exercises from the NinaPro databases, \cite{atzori2014electromyography}}
\label{fig:ninapro}
\end{figure}

For the experiments described in this paper, we utilized several publicly available databases published by the Ninapro (Non Invasive Adaptive Prosthetics) Project \cite{atzori2014electromyography}.
The aim of the project was the creation and dissemination of benchmark scientific databases of surface electromyography data for hand prostheses. The databases contain recordings from similar experiments (same number of repetitions, movements etc.), from different subjects using different recording equipment.\\
In this paper, we perform the experiments on Database 2, 3 and 4. We focus on these three because the other Ninapro databases cover more specific hardware settings and experimental situations, making them less easz to compare. The unique differences of the databases used are described in table \ref{table:db} \\
For each database, participants were asked to repeat a number of movements and force patterns. A total of 61 movement patterns (+ Rest) were used for the project and split into 4 exercise groups (see figure \ref{fig:ninapro}). Not every database covers every exercise group. We restrict our experiments to exercise group A and B, which cover basic movement of fingers and the wrist.
\\
The data acquisition procedure stays as similar as possible for every database (see figure \ref{fig:ninapro}). The participants were asked to repeat previously known movements shown via video. Each repetition took 5 seconds with a 3 second resting period in between. The number of repetitions might differ between each database. For intact subjects, the right arm was used for the exercises, while amputated subjects were asked to imagine the shown movements as naturally as possible with their missing limb.
Different kinds of electrodes were placed to record sEMG signals. Data was recorded using upto 12 electrodes, simultaneously. These electrodes were attached to the triceps brachii, biceps brachii, extensor digitorum superficialis, flexor digitorum superficialis. In addition, 8 electrodes were equally spaced around the forearm at the height of the radio-humeral joint.
For intact subjects hand kinematics were additionally measured with a CyberGlove System. 
Depending on the used sEMG-capturing system, a noise filter might be applied. All databases are published with synchronized data, super-sampled to either 2 kHz or 100 Hz using linear or nearest-neighbor interpolation.

\begin{table}[htp]
\centering
\begin{tabular}{ |c|c|c|c| } 
 \hline
 & DB2 & DB3 & DB4 \\
 Citation & \cite{atzori2014electromyography, gijsberts2014measuring} & \cite{atzori2014electromyography, atzori2016clinical} & \cite{atzori2014electromyography, pizzolato2017comparison} \\ 
 Subjects& 28 male/12 female & 11 male & 6 male/ 4 female \\ 
 Amputation? & no & yes & no \\ 
 Age & 29.9 ± 3.9 & 42.36 ± 11.96 & 29.6 ± 9.24 \\ 
 Exercise used & B (17 classes) & B (17 classes) & A (12 classes) \\ 
 Repetitions & 6 & 6 & 6 \\ 
 sEMG system & Delsys Trigno & Delsys Trigno & Cometa/Dormo \\ 
 \hline
\end{tabular}
\caption{Databases used in this paper}
\label{table:db}
\end{table}

\section*{Experiment}

\subsection*{Models and Hyperparameters}

\subsubsection*{Data Processing}
The raw data is preprocessed before the classification. Following steps are involved for preprocessing. At first, we standardize the sEMG signal and denoise it using a 4th-order butterworth bandpass filter (between 20Hz and 400Hz). Then the resulting signal is rectified and sliced using a 200 ms sliding window with 10 ms overlap. Lastly, the order of the slices is randomly shuffled, which was observed to improve generalisation in pre-tests.\\

\subsubsection*{Models}
We chose two simple deep learning models for sEMG classification.\\
The first model is a multilayer perceptron (MLP), which classifies the sEMG signal using extracted features. We choose a set of 18 commonly used sEMG features (as e.g. in \cite{pereira2022automatic}), 11 time domain and 6 frequency domain features and one wavelet based feature:\\
The time domain features are Variance, Root Mean Squared, Integral, Mean Absolute Value, Log Detector, Waveform Length, Average Amplitude Change, Difference Absolute Standard Deviation Value, Zero Crossings, Willison Amplitude and Myopulse Percentage Rate.
The frequency domain features are Frequency Ratio, Mean Frequency, Median Frequency, Peak Frequency, Mean Power, Total Power and the wavelet based feature is the Wavelet Histogram.\\
The second model is a 1-dimensional convolutional neural network (1D-CNN), which considers the preprocessed signal (2x400 (12 channels x 200ms@2kHz) ) as an input, learns the feature vectors, hierarchically and finally classify it.    \\
The motivation behind this was to enable a comparison between a feature extraction and a feature engineering approach and to see how much inter-subject variance translate to commonly used feature sets. We wanted to investigate how stable learned features are in the face of inter-subject variability and fine-tuning. We chose a one-dimensional CNN over the more common two-dimensional CNN, because we can make the results more comparable by restricting it to channel-based features.\\
This work focuses on relatively simple models compared to the existing literature (e.g. (\cite{pancholi2021robust, ding2018semg, 8641445, 8969418} ). This is a deliberate choice to keep small inference time and low computational cost. This allows us to run more experiments and also keep the option to run our models more easily on embedded hardware in the future.



\subsubsection*{Hyperparameter Tuning}
\begin{figure}[htp]
\centering
\includegraphics[width=0.8\hsize]{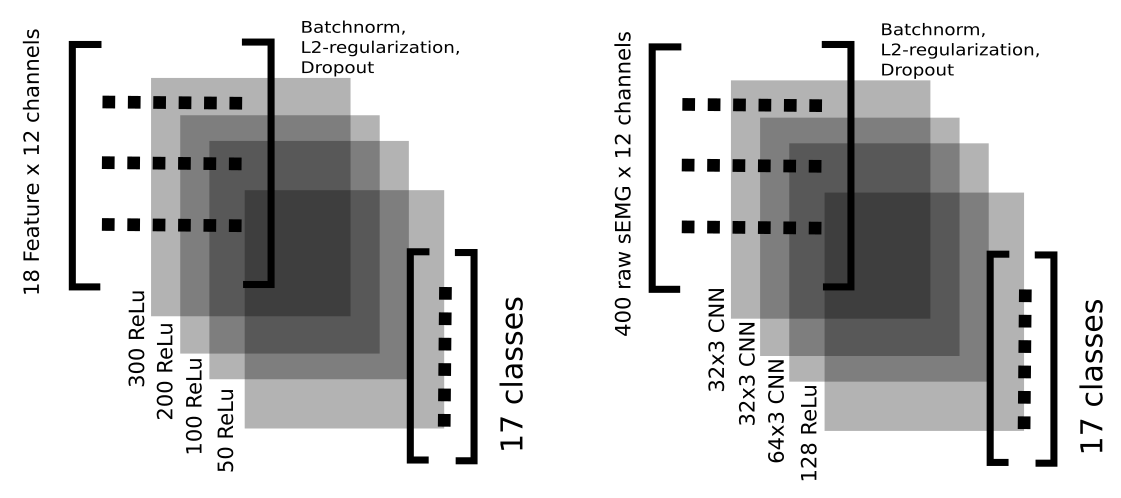}
\caption{The Models architecture found via hyperparameter tuning}
\label{fig:models}
\end{figure}

We used the first 10 subjects from DB2 to find the appropriate architecture of the employed deep learning models and tuning the parameters of the optimizer.These subjects were only used for this step and excluded from later experiments. \\
Following parameters are tuned: Number  of layers,  number of neurons of each layer, dropout factor, number of filters, filtersize, presence of batch normalization, batch size as well as the four parameters of the adam-optimizer. \\
Hyperparameter tuning was implemented using Keras \cite{chollet2015keras} and Optuna \cite{optuna_2019}.\\
The final learning parameters can be seen in table \ref{table:parameters} and the network architecture of both models is presented in figure \ref{fig:models}. 

\begin{table}[h]
\centering
\begin{tabular}{ |c|c|c|c|c|c| } 
 \hline
 Input & Learningrate & Beta1 & Beta2 & Epsilon & Batchsize \\
 Features & 0.001 & 0.4 & 0.1 & 0.38 & 256 \\ 
 Raw & 0.0002 & 0.2 & 0.9 & 0.0001 & 512 \\ 
 \hline
\end{tabular}
\caption{Hyperparameters}
\label{table:parameters}
\end{table}

\subsection*{Approach}
The aim of this paper is to evaluate the effectiveness of the  transfer learning approach for calibrating a deep learning model for a new user. To answer this question, we performed several analysis on three databases, individually.\\
For the evaluation,  we apply leave-one-subject-out cross-validation. We go through each of the N subjects in a database and individually train an artificial neural network on the remaining N-1 subjects. The task of the model is the classification of the first exercise group of each database, meaning 17 (DB2 and DB3) and 12 (DB4) classes.\\
The initial training is done on 5 out of 6 repetitions (repetition 3 is used for testing), with 10 percent of the training data used for validating the early stopping criteria. During training, we use a fixed set of hyperparameters, described in table \ref{table:parameters}.\\
During retraining the classifier on the new subjects data, a variable amount of repetitions are kept unseen for evaluation. The hyperparameters stay the same as during the initial training.\\
We compared the performance of pretrained models, before and after fine-tuning on the new subject with subject-specific models, which are only trained on data from one subject.

\section*{Evaluation}
\subsection*{DB2}
\begin{figure}[htp]
\centering
\includegraphics[width=0.8\hsize]{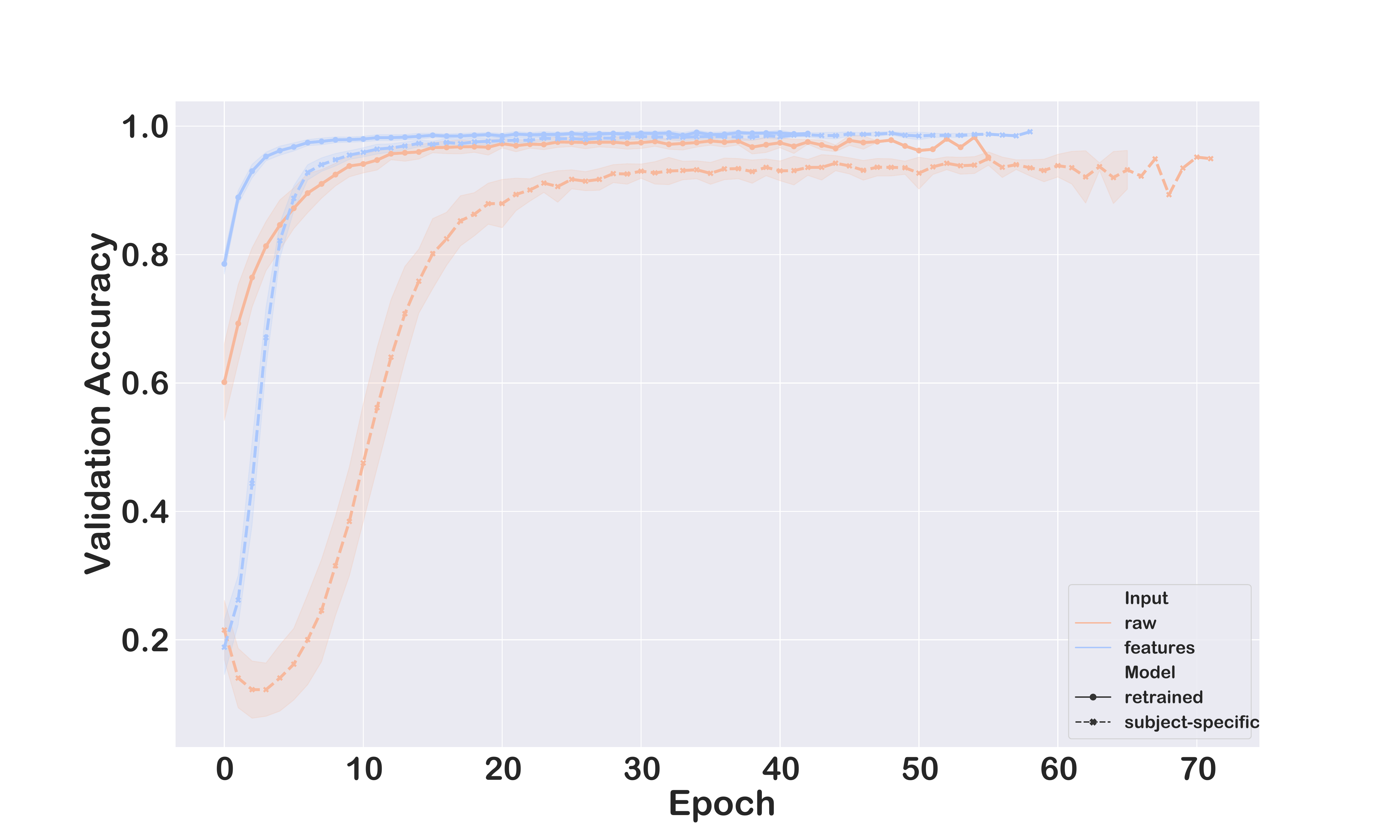}
\caption{Validation Error during training a subject-specific model vs fine-tuning the pretrained model using 5 repetitions of the new subject (DB2)}
\label{fig:db2RetrainedVSsubjectspecific}
\end{figure}

We report the evaluation performance of two models by using  transfer learning and without using transfer learning. In the case with transfer learning, we fine-tuned a pretrained model on every other subject in the dataset as described above on data from the new subject. As comparison, we also train a model with random weight initialization only on this new subjects data (subject-specific model). If transfer learning is effective for the task of sEMG classification, these subject-specific models should show a noticeable difference in their learning dynamics, compared to the models applying knowledge transfer.\\
As shown in figure \ref{fig:transfer}, signs of successful knowledge transfer are higher starting accuracy, higher slope/faster learning and/or higher asymptotic performance. We plotted the performance of subject-specific and fine-tuned models on the validation set in Figure \ref{fig:db2RetrainedVSsubjectspecific}.\\
Transferring weights from a pretrained model has a huge influence on the training dynamic for both the models, the one using features and the one using raw sEMG data as input,as shown in the figure \ref{fig:db2RetrainedVSsubjectspecific}. The validation accuracy starts at a higher point and the asymptotic accuracy is reached after a fewer number of epochs. The condition of early stopping criteria fulfilled earlier, in the case of pretrained weights. The pretrained models also reached a slightly higher final validation accuracy than the subject-specific models. \\

\begin{figure}[htp]
\centering
\includegraphics[width=0.98\hsize]{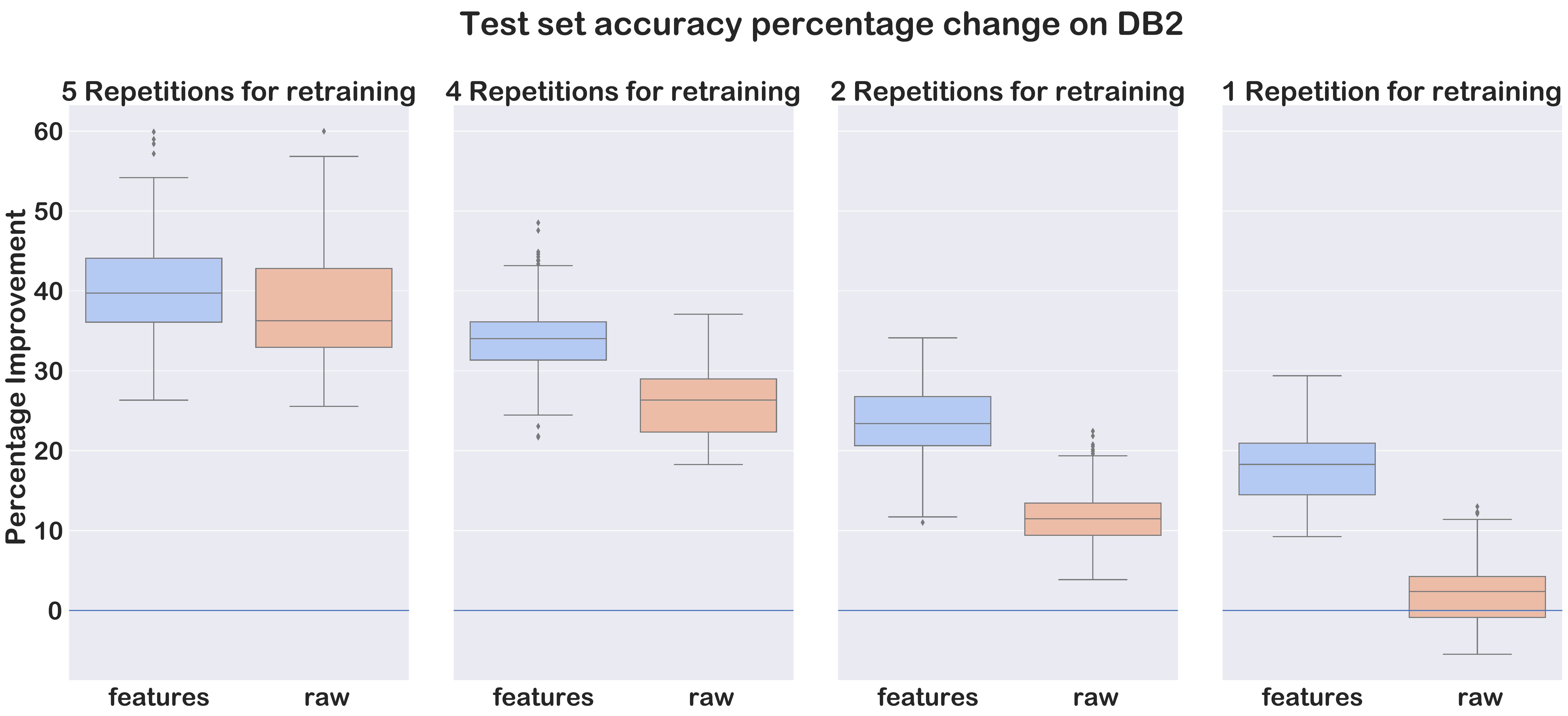}
\caption{Percentage Change of test error of the pretrained model before and after retraining on different amounts of samples (DB2)}
\label{fig:db2PercentageImprovement}
\end{figure}

While knowledge transfer seems to be taking place, we also want to evaluate the effect of fine-tuning the pretrained weights. We need to evaluate the performance on the new subjects data before and after fine-tuning to see in which situations our proposed recalibration method is successful.\\
Figure \ref{fig:db2PercentageImprovement} shows the percentage improvement due to retraining with a variable size of training data on the test data (all repetitions from the new subjects not used for training).\\
We observe a huge improvement over the pretrained model thanks to fine-tuning in situations where we have enough data from the new subject. We noticed a quick drop in percentage improvement when reducing the amount of repetitions used for fine-tuning. In the case of the model using raw input, we even notice a decrease in accuracy on the test set, which shows that fine-tuning might lead to overfitting in such situations. In general, the models using features as input, appear to be more stable and its performance degraded less quickly than the one using feature learning.\\
However, both models start to overfit when further reducing the retraining data, so that it covers less than one repetition of each movement. The simple fine-tuning approach does not seem to be appropriate for calibration with only a few selected movements.\\

\begin{figure}[htp]
\centering
\includegraphics[width=0.8\hsize]{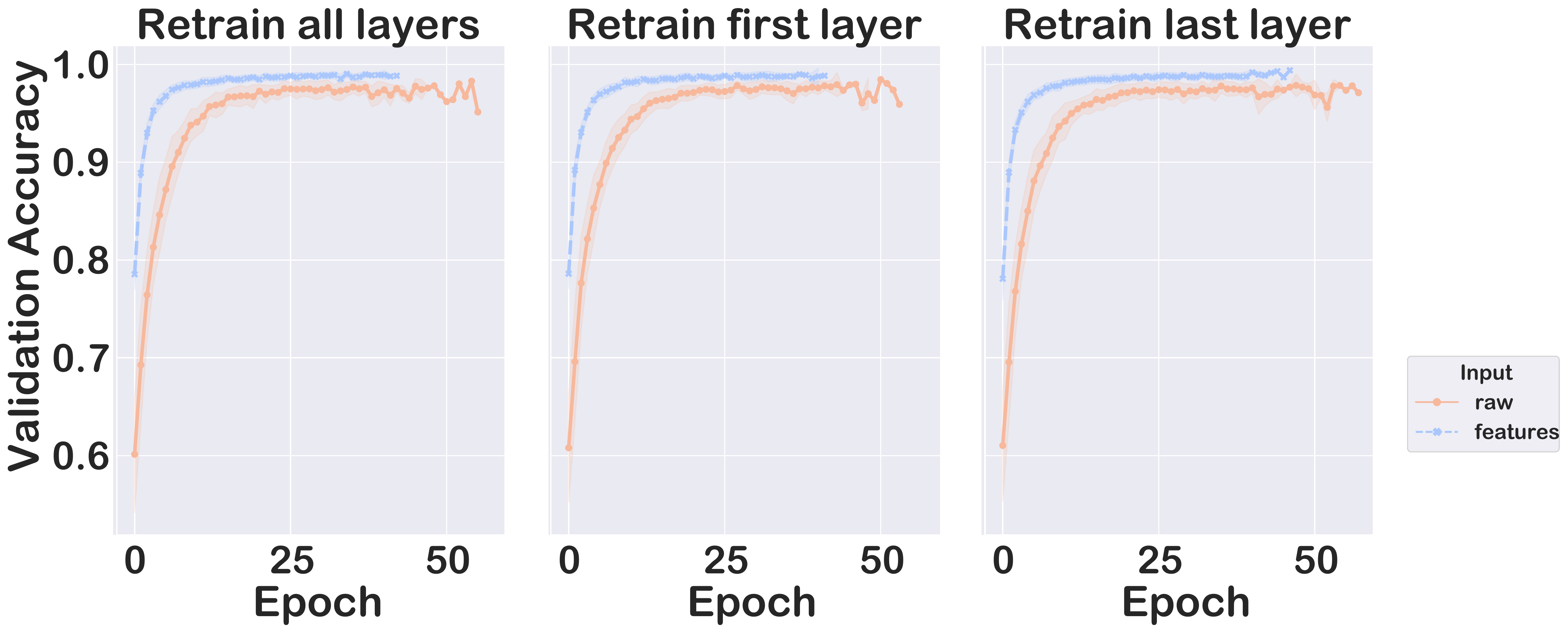}
\caption{Effect of retraining all layers or only selected layers (DB2)}
\label{fig:db2method}
\end{figure}

The most general approach to fine-tune the deep neural networks is initializing the network with the pretrained weights, and retrain all layers on the new subjects data. Often, it can be helpful to freeze parts of the network, keep the weights unchanged and use only selected layers for calibration. This might be especially the case, if we assume part of the model to be less affected by the domain change. Works like \cite{yosinski2014transferable}, show that for many common machine learning tasks, lower layers in artificial neural networks learn more general and easier to transfer features.\\
In the case of sEMG, it seems likely that it is not necessary to retrain every weight, to adopt the model to a new user.\\
We show in figure \ref{fig:db2method}, that neither of the models tested in this paper benefit from retraining all layers compared to only the first or the last layer. In practice this means that the calibration procedure can be accelerated because we can work with less tunable parameters.

Table \ref{table:db2} shows the average classification accuracy on the test repetitions generated by the new subject with a varying amount of repetitions used for retraining. As shown in figure \ref{fig:db2PercentageImprovement}, the average test accuracy of the pretrained model increases, even if just one repetition is used for fine-tuning. The feature-based, subject-specific model, outperforms the pretrained model but is again outperformed by the fine-tuned model. The same can be seen with the model using raw sEMG data, with the difference that subject specific model overfit in the low data setting and perform worse than the pretrained model.\\
Overall a sharp drop between test accuracy and validation accuracy (seen in figure \ref{fig:db2RetrainedVSsubjectspecific}) can be observed. This is due to the large in-person variance, as the validation set is a randomly taken from the repetitions used for training.\\
\begin{table}[htp]
\centering
\begin{tabular}{ccccc}
    \hline \rowcolor{hellgrau}
    \multicolumn{2}{c}{}&Subject Specific |& Before Retraining |& After Retraining\\
    \multirow{4}{*}{Features}& 1 Repetition & 0.51 & 0.49 & 0.58 \\
    & 2 Repetitions & 0.59 & 0.51 & 0.62 \\
    & 4 Repetitions & 0.65 & 0.50 & 0.67 \\
    & 5 Repetitions & 0.65 & 0.47 & 0.67 \\
    \hline
\multirow{4}{*}{raw sEMG}& 1 Repetition & 0.21 & 0.56 & 0.58 \\
    & 2 Repetitions & 0.23 & 0.57 & 0.64\\
    & 4 Repetitions & 0.51 & 0.54 & 0.68 \\
    & 5 Repetitions & 0.64 & 0.49 & 0.68\\
    \hline
\end{tabular}

\caption{Average classification accuracy on unseen test data from new subject (DB2)}
\label{table:db2}
\end{table}

\subsection*{DB3}
\begin{figure}[htp]
\centering
\includegraphics[width=0.8\hsize]{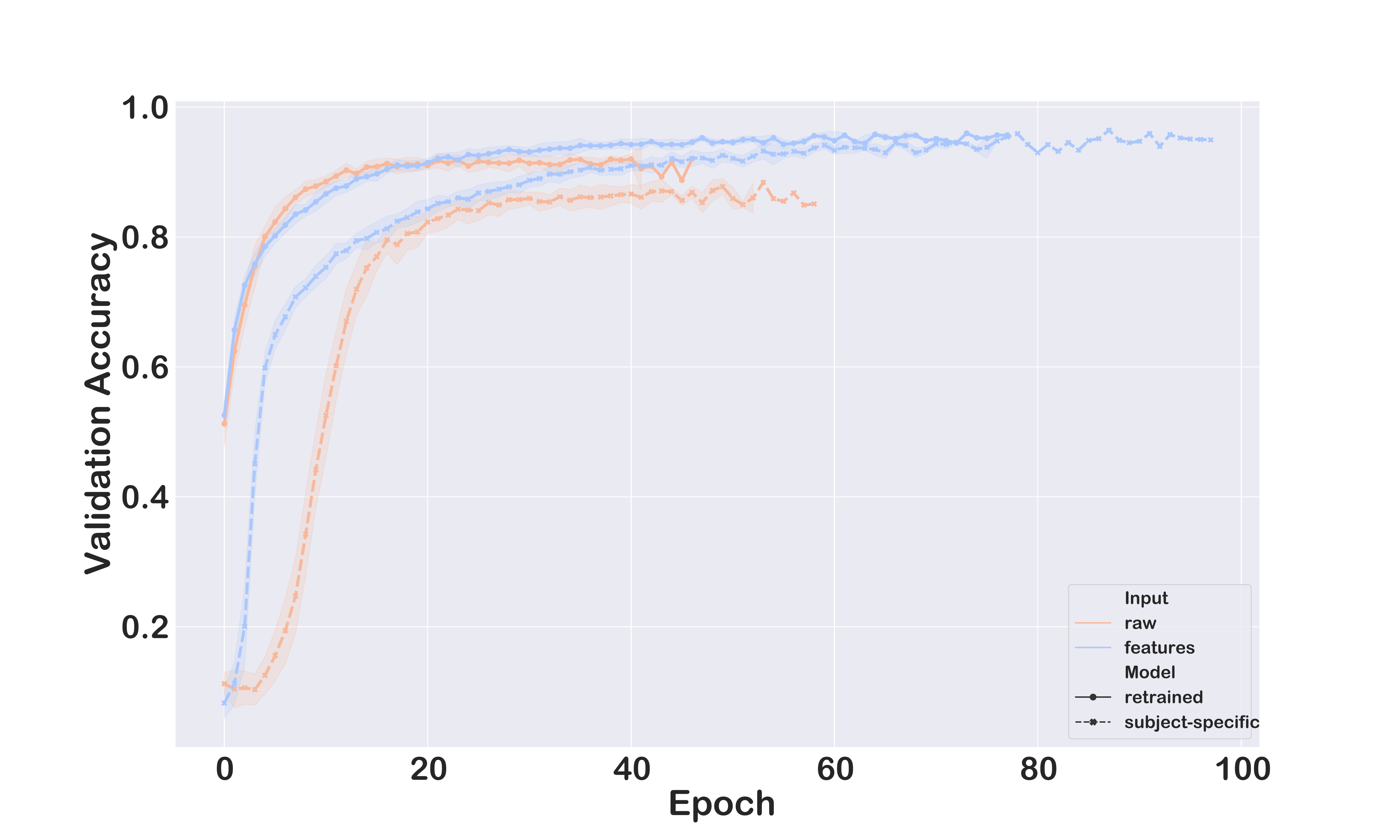}
\caption{Validation Error during training a subjectspecific model vs fine-tuning the pretrained model (DB3)}
\label{fig:db3RetrainedVSsubjectspecific}
\end{figure}

\begin{figure}
\centering
\includegraphics[width=0.98\hsize]{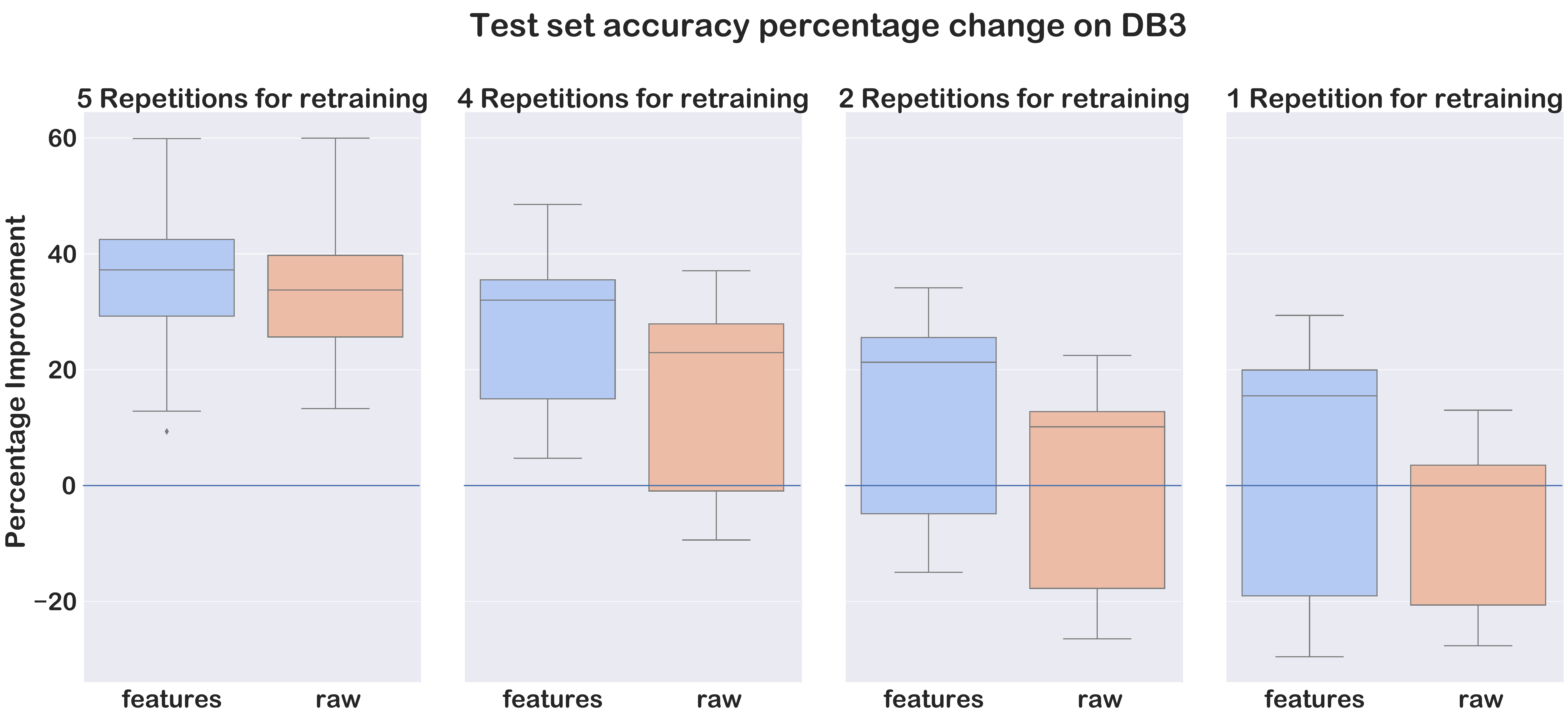}
\caption{Percentage Change of test error of the pretrained model before and after retraining on different amounts of samples (DB3)}
\label{fig:db3PercentageImprovement}
\end{figure}

\begin{figure}
\centering
\includegraphics[width=0.8\hsize]{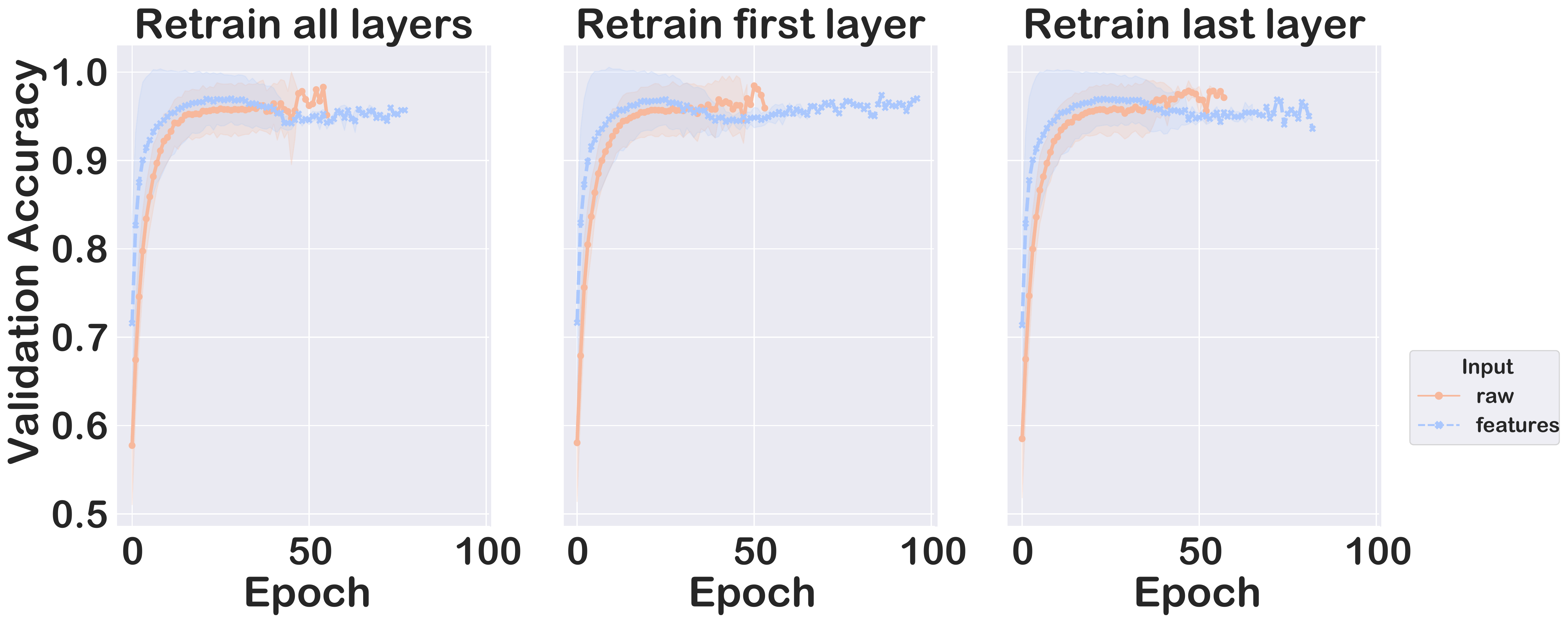}
\caption{Effect of retraining all layers or only selected layers (DB3)}
\label{fig:db3method}
\end{figure}

The data acquisition setup for database 3 was identical to database 2. However, using a smaller number of total subjects, which where all amputated.\\ 
We expected to observe a higher variability between subjects, which may result in overall poor performance. However, a stronger impact of fine-tuning is expected.\\
 Table \ref{table:db3} shows that overall performance of the classifiers is indeed degraded. However, in leave-one-subject-out cross-validation low variability is observed as well as a lower percentage improvement between the pretrained and the fine-tuned model.
This might be explained with the amputation having a lower impact on between-subject variability than expected. Other factor such as sex, which are more diverse in database 2, could have an higher influence.\\
Despite this, figures \ref{fig:db3method},\ref{fig:db3PercentageImprovement} and \ref{fig:db3RetrainedVSsubjectspecific} show similar trends as already observed on database 2.\\
Again, we observe quicker learning and higher asymptotic performance compared to the subject-specific models and we see no noticeable difference between retraining different layers. The improvement due to fine-tuning drops even faster, when reducing the amount of repetitions, with the feature-based models being more stable in low-data situations.\\
Why the models start earlier to overfit is difficult to explain. One possibility might be that the base pretrained model is unable to generalize well. This could be caused by the lower number of subjects used during pre-training, or due to the values of  hyperparameters that were kept fixed (taken from database 2).


\begin{table}[htp]
\centering
\begin{tabular}{ccccc}
    \hline   \rowcolor{hellgrau}
    \multicolumn{2}{c}{}&Subject Specific |& Before Retraining |& After Retraining\\
    \multirow{4}{*}{Features}& 1 Repetition & 0.33 & 0.46 & 0.36 \\
    & 2 Repetitions & 0.41 & 0.47 & 0.42 \\
    & 4 Repetitions & 0.46 & 0.43 & 0.47 \\
    & 5 Repetitions & 0.50 & 0.41 & 0.50 \\
    \hline
\multirow{4}{*}{raw sEMG}& 1 Repetition & 0.15 & 0.59 & 0.45 \\
    & 2 Repetitions & 0.16 & 0.58 & 0.45\\
    & 4 Repetitions & 0.48 & 0.54 & 0.51 \\
    & 5 Repetitions & 0.52 & 0.43 & 0.52\\
    \hline
\end{tabular}
\caption{Average classification accuracy on unseen test data from new subject (DB3)}
\label{table:db3}
\end{table}

\subsection*{DB4}
\begin{figure}[htp]
\centering
\includegraphics[width=0.8\hsize]{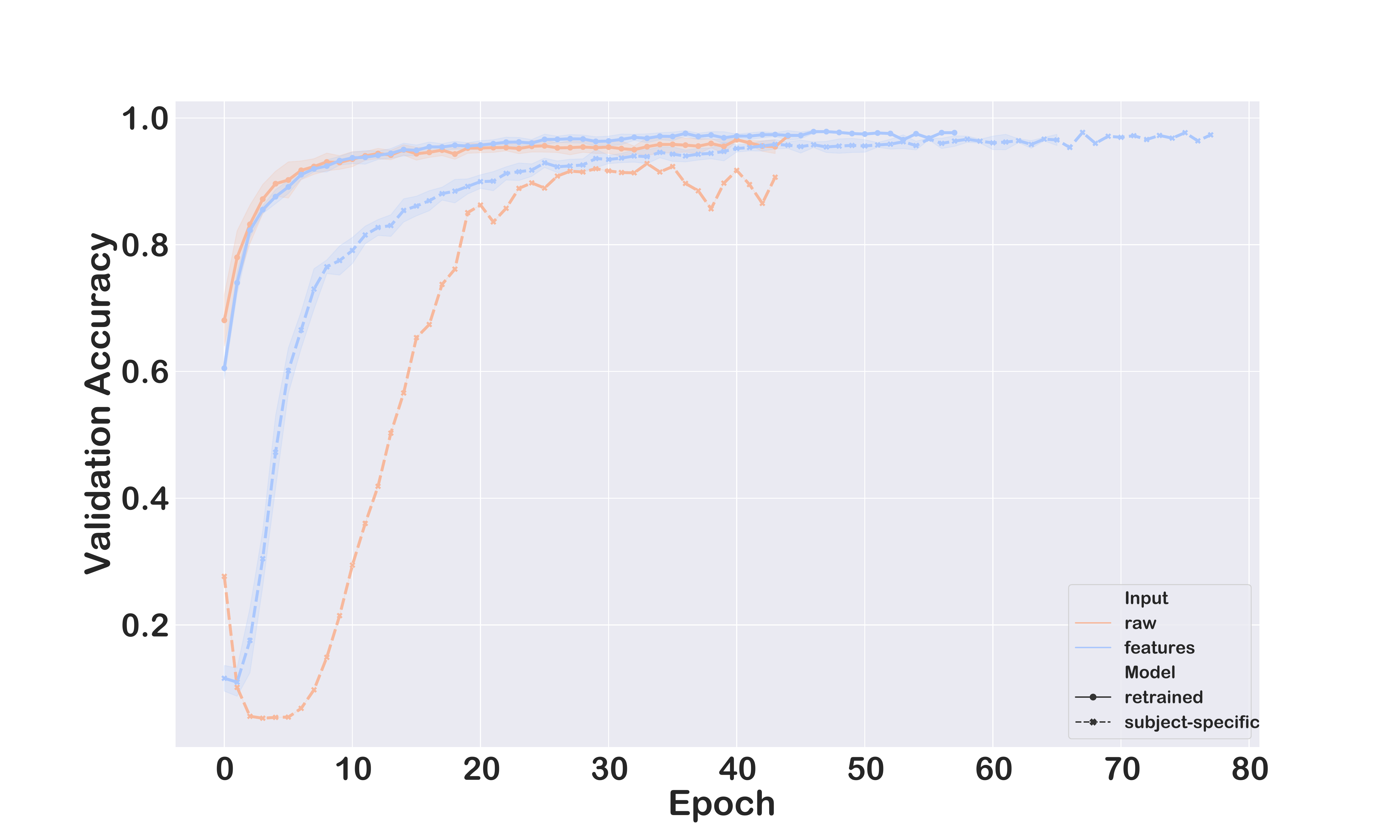}
\caption{Validation Error during training a subject specific model vs fine-tuning the pretrained model (DB4)}
\label{fig:db4RetrainedVSsubjectspecific}
\end{figure}

\begin{figure}[htp]
\centering
\includegraphics[width=0.98\hsize]{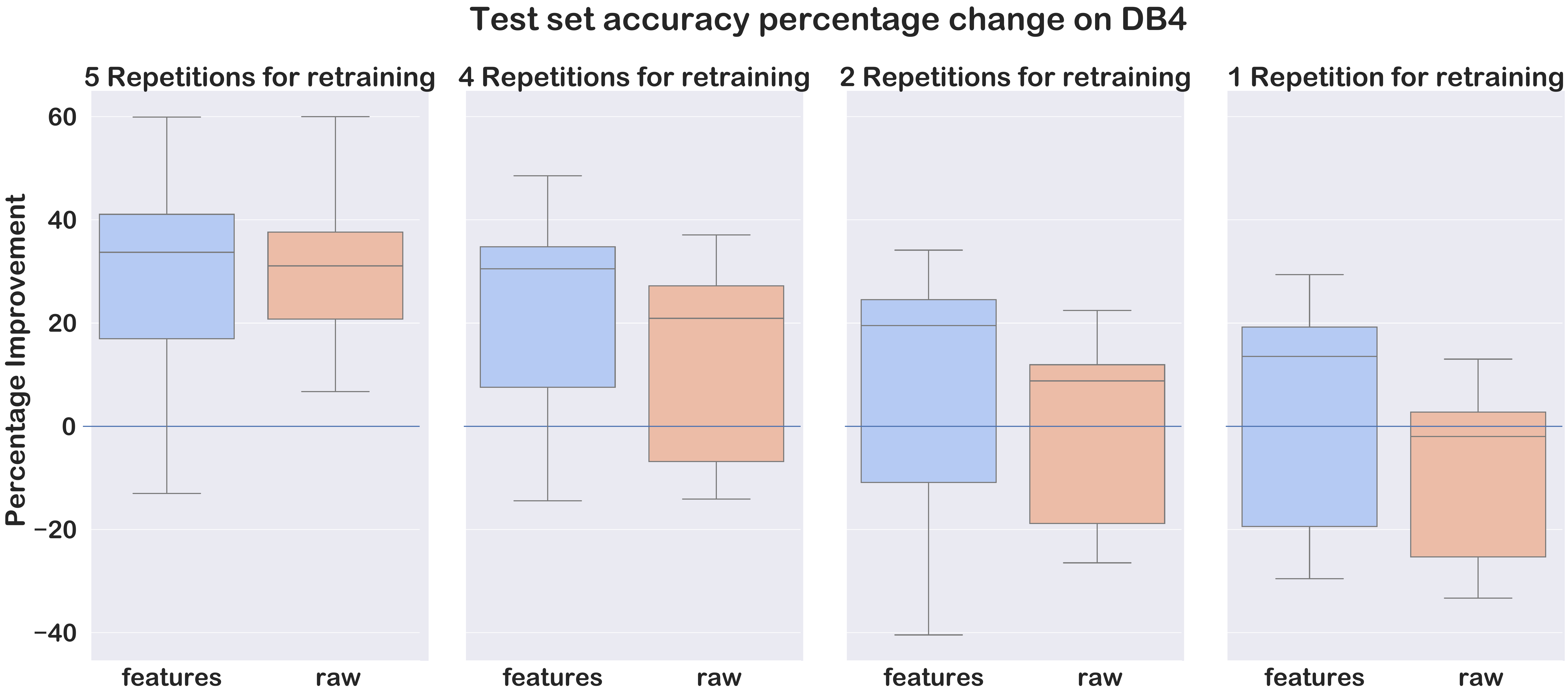}
\caption{Percentage Change of test error of the pretrained model before and after retraining on different amounts of samples (DB4)}
\label{fig:db4PercentageImprovement}
\end{figure}

\begin{figure}
\centering
\includegraphics[width=0.8\hsize]{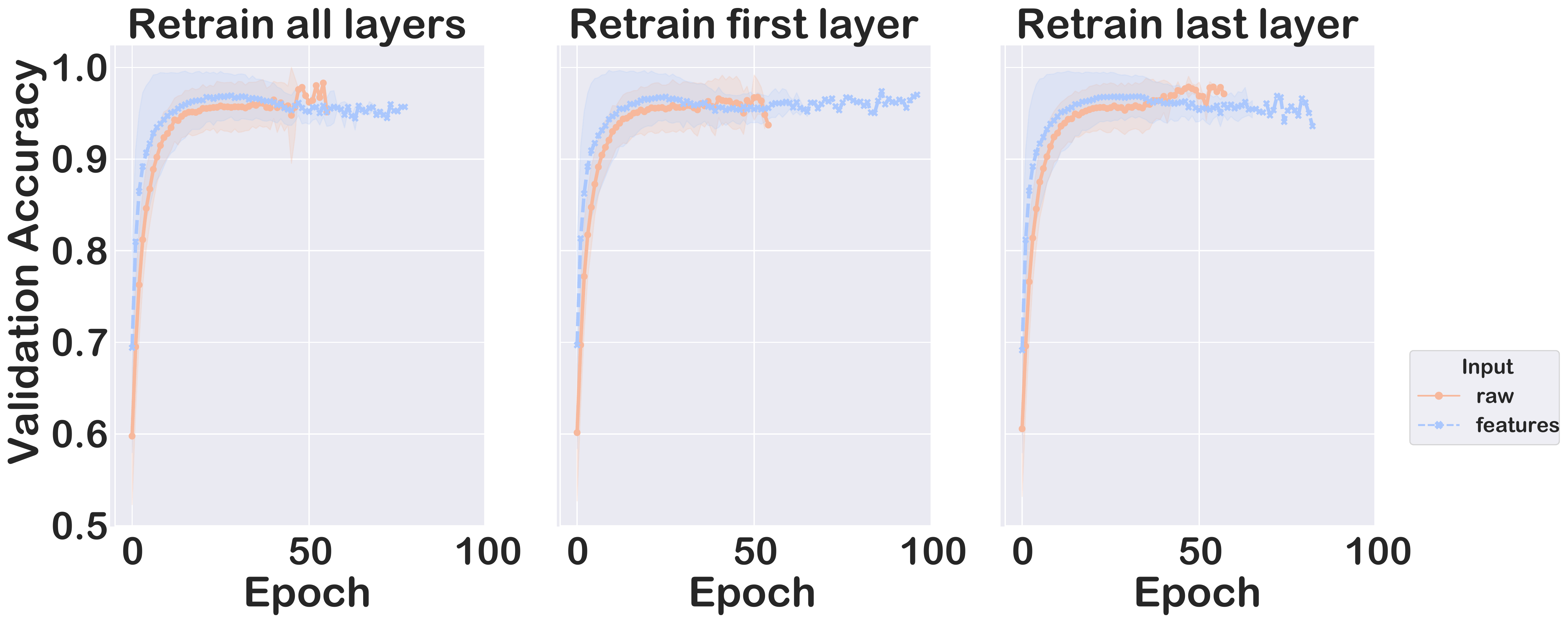}
\caption{Effect of retraining all layers or only selected layers (DB4)}
\label{fig:db4method}
\end{figure}

We observe similar training curves (figure \ref{fig:db4RetrainedVSsubjectspecific} and effects of retraining method (figure \ref{fig:db4method}) with database 4, as in the previous two databases. Here, the subject-specific models perform again worse than the pretrained models.\\
Looking at the percentage improvement (figure \ref{fig:db4PercentageImprovement}) and overall performance (table \ref{table:db4}) , we notice a different pattern than before.\\
On database 4, the feature learning (raw) model does not appear to be less stable than the one using manually extracted features. This is the only case where the model does not overfit using raw sEMG inputs with five repetitions for training.\\
A possible reason for this behaviour might be the different movement tasks compared to DB2 and 3. Previous databases(DB2 and 3)  used various hand configurations and wrist movements, while DB4 uses finger movements. It seems plausible that this task benefits stronger from being able to learn features because the activity measured takes place further away from the sEMG-electrode locations and might therefore be less well captured by feature engineering approach.\\
We however also see in table \ref{table:db4}, that the before retraining performance varies more strongly over the experimental settings than on the other two databases. This suggests that the intra-subject variability between each repetition might be higher than in those other databases. One possible cause for this could be the different movement tasks and recording equipment.   
The overall lower accuracy (compared to database 2) might be due to hyperparameters not being exactly tuned for this database and the lower number of subjects.


\begin{table}[htp]
\centering
\begin{tabular}{ccccc}
    \hline \rowcolor{hellgrau}
    \multicolumn{2}{c}{}&Subject Specific |& Before Retraining |& After Retraining\\
    \multirow{4}{*}{Features}& 1 Repetition & 0.39 & 0.55 & 0.45 \\
    & 2 Repetitions & 0.34 & 0.55 & 0.37 \\
    & 4 Repetitions & 0.45 & 0.51 & 0.47 \\
    & 5 Repetitions & 0.45 & 0.46 & 0.46 \\
    \hline
\multirow{4}{*}{raw sEMG}& 1 Repetition & 0.22 & 0.71 & 0.50 \\
    & 2 Repetitions & 0.29 & 0.69 & 0.57\\
    & 4 Repetitions & 0.38 & 0.63 & 0.55 \\
    & 5 Repetitions & 0.50 & 0.47 & 0.54\\
    \hline
\end{tabular}

\caption{Average classification accuracy on unseen test data from new subject (DB4)}
\label{table:db4}
\end{table}

\subsection*{Comparison with other works on NinaPro}
Comparing our models with other works on the same database is difficult, due to differences in the chosen exercise group, input features, window size and the unbalanced nature of the NinaPro databases (e.g. in \cite{josephs2020semg} a drop of more than ten percentage points was noticed due to balancing issues in database 5).\\
However, when comparing the accuracy achieved by models in this work with previously achieved performance (see table \ref{table:OtherPapersClassification}), our models lie on the lower end of the spectrum. This is not surprising due to our focus on simpler models with fewer tunable parameters. Looking at the only other paper working with the same simple 1D-CNN, our raw model performs comparatively well.\\
We don't have reasons to assume that applying larger models would affect the general tendencies we found in regard to the effects of transfer learning. Increasing the parameter space would likely only lead to an even wider gap between subject-specific models and pretrained models, while having a better performing base model, might further strengthen the regularization provided by transfer learning. In pre-tests performed using 2D-CNNs, we didn’t notice any effects of transfer learning that deviate from the effects presented in this paper.\\

 \renewcommand{\arraystretch}{1.15}
\begin{table}[htp]
\centering
\resizebox{\textwidth}{!}{
\begin{tabular}{ccccccccc}
    \hline \rowcolor{hellgrau}
    Citation & Database    &Classes&Balanced&Input &Window    &Model&    Subject-specific&    Result\\
    \hline
    \cite{8641445} &    DB2    &50&    no&    sEMG features & 200 ms    &    Multiview 2D-CNN& yes&    83.7\\
    \hline
    
    \multirow{4}{*}{\cite{pancholi2021robust}}&
    \multirow{2}{*}{\parbox{2.5cm}{\centering DB2 \\ (10 subjects)}}& 
    \multirow{4}{*}{49}&
    \multirow{4}{*}{yes}&
    spectrogram&
    \multirow{4}{*}{150 ms}&
    \multirow{4}{*}{2D-CNN}&
    \multirow{4}{*}{yes?}&
    73.12\\
    &&&&MPP+MZP&&&&89.45\\
    &\multirow{2}{*}{\parbox{2.5cm}{\centering DB3 \\ (5 subjects)}}&&&spectrogram&&&&66.31\\
    &&&&MPP+MZP&&&&81.67\\
    \hline
    
    \multirow{2}{*}{\cite{xiemovement}}&
    \multirow{2}{*}{DB 2}&
    \multirow{2}{*}{18}&
    \multirow{2}{*}{no}&
    \multirow{2}{*}{raw sEMG}&
    \multirow{2}{*}{64 ms}&
    1D-CNN&
    \multirow{2}{*}{no?}&
    52.17\\
    &&&&&&1D-CNN+RNN&&63.74\\
    \hline
    
    \multirow{2}{*}{\cite{10.3389/fnins.2017.00379}}&
    \multirow{2}{*}{DB 2}&
    \multirow{2}{*}{17}&
    \multirow{2}{*}{yes}&
    \multirow{2}{*}{raw sEMG}&
    \multirow{2}{*}{200 ms}&
    \multirow{2}{*}{2D-CNN}&
    \multirow{2}{*}{yes?}&
    82.22\\
    &&49&&&&&&78.71\\
    \hline

    \cite{8969418} & DB2 & 17 &    yes    & raw sEMG &    100 ms &    Dilated Causal CNN & no? &    92.5\\
    \hline
    
    \multirow{2}{*}{\cite{ding2018semg}}&
    \multirow{2}{*}{DB 2}&
    \multirow{2}{*}{17}&
    \multirow{2}{*}{yes}&
    \multirow{2}{*}{raw sEMG}&
    \multirow{2}{*}{100 ms}&
    \multirow{2}{*}{2-Block 2D-CNN} &
    \multirow{2}{*}{yes?}&
    83.17\\
    &&49&&&&&&78.66\\
    \hline

\end{tabular}
}
\caption{Classification accuracies achieved on NinaPro Databases using Deep Learning}
\label{table:OtherPapersClassification}
\end{table}

The few other papers working on deep transfer learning on NinaPro data can be seen in table \ref{table:OtherPapersTransfer}. \\
On Database 2, our models outperforms \cite{9495836} in base classification accuracy, in settings where we can retrain on many repetitions, and percentage improvement in any tested setting. On the other two databases, their models outperformed those tested in this paper, which again might be due to our hyperparameters being tuned on database 2. Their distribution alignment approach has the additional advantage, that it only needs unlabeled data for retraining.\\
The approach in \cite{ketyko2019domain} is quite similar to ours, as it also uses a weight-initialization approach. Here however, they freeze all pretrained layers and add a new 'adaptation layer'. It is not apparent how either of the two approaches to fine-tuning should differ in principle. Our results are difficult to compare because they use database 1, which differs a lot from the databases used in our work. They report a lower base performance but a higher percentage improvement using 5 repetitions.\\
\begin{table}[htp]
\centering
\begin{tabular}{cccccccc}
    \hline \rowcolor{hellgrau}
    Citation & Database    &Classes&Balanced&Method&mMdel&without TL&with TL\\
    \hline
    
    \multirow{3}{*}{\cite{9495836}}&
    DB2& 
    \multirow{2}{*}{50}&
    \multirow{3}{*}{no}&
    \multirow{3}{*}{distribution alignment + DNM}&
    \multirow{3}{*}{(TL-)MKCNN}&
    63.73&
    65.29\\
    &DB3&&&&&76.84&82.47\\
    &DB4&53&&&&60.10&62.33\\
    \hline
    
    \cite{s17030458} &    DB1 &    52 &yes    &    AdaBN &    ensemble CNN    & - &    67.4\\
    \hline
    
    \cite{ketyko2019domain} &DB1 &     12    &yes &    domain adaptation layer &    LSTM &    35.10    & 65.29\\
    \hline

\end{tabular}
\caption{Other Deep Transfer Learning results NinaPro Databases }
\label{table:OtherPapersTransfer}
\end{table}

\newpage

\section*{Conclusion}
The  conducted experiments indicate that relying on training user-specific deep learning models for sEMG classification is not advisable. Fine-tuning of a pretrained model on a new user data did outperform the user-specific trained models in every setting covered by this paper. This is likely caused by additional regularization provided through the weight-initialization based transfer learning. Given the right circumstances, even this relatively simple approach to transfer learning seems to be able to perform comparatively well in improving the classification accuracy on sEMG over subject-specific models.\\
Being able to reduce the parameter size by freezing all but the last layer and stopping after a few epochs of retraining, makes this a feasible calibration method in practical settings.
While it would be desirable to calibrate on just a few data points, reducing the number of iterations used for retraining can lead quickly to overfitting to the given samples. It should be noted that we didn't change the hyperparameters for the fine-tuning stage. So it is possible that a more optimal set of learning parameters could lead to a better performance in low data situations.\\
The better performance on the larger database 2 indicates that having a better performing base model, pretrained on more subjects might lead to a stronger regularization and a more stable behaviour during retraining. It is however difficult to know upfront which model could work as a stable basis, as preretraining performance does not appear to aid as a clear indicator for successful fine tuning. While both the models using raw data and the model using manual features showed a similar classification accuracy, the latter showed more stability during fine-tuning and was less prone to overfitting on fewer data points.  Following from this, potential future research could investigate what features of a model make it a promising base model for further fine-tuning. \\
Another open question is the effect of intra-personal variability of the sEMG signal. This work focused on interpersonal variability and model transfer between subjects. However as we saw, especially in database 4, the same model showed highly different base performance depending on which repetition of the same person the samples were coming from. This takes into question whether a one-time calibration on a few data points gathered from a new user might ever be enough to generate a stable model that accounts for the variability of sEMG signal.  Future work should focus on whether inter- or intra-personal variance of sEMG signals have a stronger influence on deep learning based classifiers. These results could help find appropriate calibration schemes for sEMG classifiers in practical settings.

\bibliography{sample}

\section*{Acknowledgements}

This work is supported by the Ministry of Economics, Innovation, Digitization and Energy of the State of North Rhine-Westphalia and the European Union, grants GE-2-2-023A (REXO) and IT-2-2-023 (VAFES)

\section*{Author contributions statement}

\textbf{Conceptualization:} SL MS TG II.\\
\textbf{Funding acquisition:} II.\\
\textbf{Investigation:} SL.\\
\textbf{Writing – original draft:} SL\\
\textbf{Writing – review\& editing:} SL MS TG II


\section*{Corresponding author}
Correspondence to Stephan Lehmler

\section*{Competing interests statement}
The authors declare no competing interests.

\end{document}